\documentclass[sigconf,natbib=true,anonymous=false]{acmart}

\AtBeginDocument{%
  \providecommand\BibTeX{{%
    \normalfont B\kern-0.5em{\scshape i\kern-0.25em b}\kern-0.8em\TeX}}}

\copyrightyear{2021} 
\acmYear{2021} 
\setcopyright{rightsretained} 
\acmConference[CIKM '21]{Proceedings of the 30th ACM International Conference on Information and Knowledge Management}{November 1--5, 2021}{Virtual Event, QLD, Australia}
\acmBooktitle{Proceedings of the 30th ACM International Conference on Information and Knowledge Management (CIKM '21), November 1--5, 2021, Virtual Event, QLD, Australia}
\acmDOI{10.1145/3459637.3482488}
\acmISBN{978-1-4503-8446-9/21/11}

\usepackage{amsmath}
\usepackage{amssymb}
\usepackage{amsfonts}
\usepackage{booktabs}
\usepackage{algorithmic}
\usepackage{array}
\usepackage{multirow}
\usepackage{stfloats, midfloat}
\usepackage{color, graphics}
\usepackage{multirow}
\usepackage{xspace}
\usepackage{caption}
\usepackage{balance} 
\usepackage{textcomp}
\usepackage{makecell}
\usepackage{wasysym}

\newcommand{\nodenorm}{$\mathrm{NodeNorm}$\xspace}

\newcommand{\nodenormp}{$\mathrm{NodeNorm}_p$\xspace}
\newcommand{\nodenormone}{$\mathrm{NodeNorm}_1$\xspace}
\newcommand{\nodenormtwo}{$\mathrm{NodeNorm}_2$\xspace}
\newcommand{\nodenormthree}{$\mathrm{NodeNorm}_3$\xspace}

\newcommand{\prop}{\texttt{PROP}\xspace}
\newcommand{\props}{\texttt{PROPs}\xspace}
\newcommand{\tran}{\texttt{TRAN}\xspace}
\newcommand{\trans}{\texttt{TRANs}\xspace}
\newcommand{\varianceinflaming}{\textit{variance inflammation}\xspace}
\definecolor{mycolor}{RGB}{219,90,107}

\newcommand{\gray}[1]{\textcolor{gray}{#1}}

\makeatletter
\DeclareRobustCommand\onedot{\futurelet\@let@token\@onedot}
\def\@onedot{\ifx\@let@token.\else.\null\fi\xspace}

\def\eg{\emph{e.g}\onedot} \def\Eg{\emph{E.g}\onedot}
\def\ie{\emph{i.e}\onedot}

\def\wrt{w.r.t\onedot} 
\def\etal{\emph{et al}\onedot}
\makeatother

\setcopyright{rightsretained} 
\begin{document}
\fancyhead{}

\title{Understanding and Resolving Performance Degradation in Deep Graph Convolutional Networks}

\author{
	Kuangqi Zhou$^{1*}$, Yanfei Dong$^{1,2*}$ \\Kaixin Wang $^1$, Wee Sun Lee$^1$,
  Bryan Hooi$^1$,
  Huan Xu$^3$,
  Jiashi Feng$^1$
}
\thanks{* Equal Contribution.}
\affiliation{
	 $^1$National University of Singapore \hspace{5mm}
	 $^2$PayPal Innovation Lab \hspace{5mm}
	 $^3$Alibaba Group \\
	 \{kzhou, dyanfei, kaixin.wang\}@u.nus.edu \hspace{2mm} \{dcsleews, dcsbhk, elefjia\}@nus.edu.sg \hspace{2mm}  huan.xu@alibaba-inc.com %
	 \country{}
}





  



\begin{abstract}
A Graph Convolutional Network (GCN) stacks several layers and in each layer performs a PROPagation operation~(\prop) and a TRANsformation operation~(\tran) for learning node representations over graph-structured data. Though powerful, GCNs tend to suffer performance drop when the model gets deep. Previous works focus on \props to study and mitigate this issue, but the role of \trans is barely investigated.
In this work, we study performance degradation of GCNs by experimentally examining how stacking only \trans or \props works.
We find that  \textit{\trans contribute significantly, or even more than \props}, to declining performance, and moreover that they tend to amplify node-wise feature variance in GCNs, causing \varianceinflaming that we identify as a key factor for causing performance drop.
Motivated by such observations, we propose a variance-controlling technique termed Node Normalization (\nodenorm), which scales each node's features using its own standard deviation.
Experimental results validate the effectiveness of \nodenorm on addressing performance degradation of GCNs.
Specifically, it enables deep GCNs to outperform shallow ones in cases where deep models are needed, and to achieve comparable results with shallow ones on 6 benchmark datasets.
\nodenorm is a generic plug-in and can well generalize to other GNN architectures. Code 
is publicly available at \url{https://github.com/miafei/NodeNorm}.
\end{abstract}




\maketitle

\section{Introduction}\label{sec:introduction}
\label{introduction}
Graph Neural Networks~(GNNs)~\cite{kipf2016semi, hamilton2017inductive, velivckovic2017graph, xu2018representation} have been proposed and widely adopted to learn from graph-structured data. GNNs stack several layers, and in each layer perform a \textit{PROPagation operation}~(\prop) and a \textit{TRANsformation operation}~(\tran)~\cite{zhou2018graph} to produce informative node representations that can be used to facilitate downstream tasks like node classification~\cite{zhou2018graph, hu2020ogb, shchur2018pitfalls}. 

Though achieving remarkable success, GNNs suffer a model depth limitation\textemdash they tend to perform increasingly worse on classifying graph nodes as the model gets deeper~\cite{li2018deeper, chen2019measuring, li2019deepgcns}.
This performance degradation problem has been widely explored in previous literature~\cite{li2018deeper, li2019deepgcns, chen2019measuring, Rong2020DropEdge:, zhao2019pairnorm}, but with the focus on effects of \props operations of the Graph Convolutional Networks~(GCNs).
GCNs are a representative GNN architecture that uses the 1-order approximation of the Chebyshev polynomials of the graph Laplacian matrix~\cite{defferrard2016convolutional} to perform graph convolution~\cite{kipf2016semi, zhou2018graph}.
Recent works~\cite{li2018deeper, chen2019measuring, zhao2019pairnorm} observe that \props in deep GCNs overly mix the hidden features of different nodes, a.k.a. \emph{oversmoothing}, and hence hurt the node classification performance.
Moreover, \props are also believed to cause \emph{gradient vanishing}, thus hindering the model training process and leading to poor performance~\cite{li2019deepgcns, Rong2020DropEdge:}.

However, the role of \trans, the other major operation of a GCN layer, in this problem is largely neglected and less understood. 
In this work, we investigate on the effects of \trans upon the performance of GCNs \wrt model depth, to provide a more comprehensive understanding of the reasons for the performance degradation problem of GCNs.
In particular, we carefully design a set of ablative experiments where we disentangle the \props and \trans to check their respective effects on model performance. 
We design two variants of GCNs: one only performs \trans in hidden layers, and the other only performs \props in hidden layers for learning the node representations. 
We observe that the former model, \ie performing only \trans, generally causes \textit{more} significant performance drop than performing only \props. 
This surprising finding differs from the previous common belief that oversmoothing and gradient vanishing caused by \props are the main reasons for performance degradation.
Actually, \textit{\trans contribute significantly, or even more}.

Such intriguing observations motivate us to dig deeper into the reasons behind performance degradation. We then investigate how \trans hurt the performance by examining their influence on the node representations. 
We observe that \trans tend to amplify node-wise feature variance (\ie, the variance of each node's hidden features). 
Consequently, as a GCN gets deeper, the number of \trans increases and the node-wise feature variance in general increases greatly. 
We refer to this phenomenon as \varianceinflaming.  
Moreover, we find that larger variance of node features leads to greater difficulty in classifying these nodes, and thus deep GCNs perform significantly worse than shallow ones.
The effects of large feature variance on classification performance remind us of some early studies on Multi-Layer Perceptrons~(MLPs) and Convolution Neural Networks~(CNNs)~\cite{klambauer2017self, ioffe2015batch, ba2016layer}. 
Though those works similarly claim that the large feature variance would affect model performance, they do not establish the explicit relation to the difficulty of training a deeper model.
Moreover, due to the complex entanglement between the \props and \trans in GCNs, such a factor of large feature variance is hidden by the feature smoothing phenomenon among nodes (\ie, oversmoothing). 

We are motivated to mitigate \varianceinflaming to address the performance degradation issue for deep GCNs.
To this end, we propose a plug-in variance-controlling technique termed as Node Normalization~(\nodenorm) that can effectively alleviate \varianceinflaming. 
\nodenorm scales hidden features of each single node based on its standard deviation. 
To make the normalization effect controllable and collaborate well with GCNs in different scenarios, \nodenorm takes the $p$-th root of the standard deviation as the normalization factor. 
With a smaller $p$, \nodenorm controls node-wise feature variance more strictly. 
In the following part of our work,
we use \nodenormp to denote \nodenorm with a specific $p$.
We empirically find \nodenorm is effective for improving the performance of deep GCNs by well handling the \varianceinflaming problem. 
To further reveal the importance of variance controlling, we also investigate whether and why the existing Layer Normalization~(LayerNorm)~\cite{ba2016layer}, which also performs a node-wise variance-scaling operation, helps reduce performance degradation. We find through experiments that LayerNorm also mitigates \varianceinflaming thus improving deep model performance.

Extensive experiments on various types of graph datasets demonstrate that mitigating \varianceinflaming successfully relieves the performance degradation of deep GCNs.
 Specifically, we make following observations: 1)~The \nodenorm enables deep GCNs (\eg 64-layer) to \emph{outperform} shallow ones (\eg 2-layer) in 3 cases~\cite{zhao2019pairnorm, sun2019multi, jia2020residual} where usually deep models are required to learn good node representations. 2)~With \nodenorm, deep GCNs (\eg 64-layer) can achieve results comparable to shallow ones (\eg 2-layer) on 6 benchmark datasets~\cite{sen2008collective, shchur2018pitfalls, mernyei2020wiki, shchur2018pitfalls}; 3)~The \nodenorm outperforms two existing best-performing methods~\cite{zhao2019pairnorm, Rong2020DropEdge:} for addressing performance degradation in deep GCNs in most cases.
 4)~We reveal that the true contributing factor of LayerNorm's success in improving deep model performance is its variance scaling step. 

Our proposed \nodenorm is generic and generalizable to other GNNs. Specifically, \nodenorm resolves performance degradation for popular GNNs including GAT~\cite{velivckovic2017graph} and GraphSage~\cite{hamilton2017inductive}. It also improves performance of recent deep GNN architectures that do not suffer performance degradation, including GCNII~\cite{chen2020simple} and GEN~\cite{li2020deepergcn}.

The contributions of this paper are three-fold: 
\begin{itemize}
 \setlength\itemsep{0em}
\item We empirically find that \trans make a significant cause of the performance degradation problem for deep GCNs, which is however under-explored in previous works.

\item We figure out through experiments that \trans cause \varianceinflaming, and that deep GCNs perform notably worse on nodes with relatively large variance as compared to shallow models. 
Based on these findings, we propose \nodenorm to mitigate \varianceinflaming. 

\item The proposed \nodenorm well resolves performance degradation of GCNs, and enables deep GCNs (\eg 64-layer) to outperform shallow ones in cases where often deep models are required to learn good node representations. Moreover, it helps deep GCNs to achieve comparable results with shallow ones on benchmark datasets.
\end{itemize}

\section{Related works}
\label{sec:related-works}
\subsection{Graph Neural Networks (GNNs)}
Graph neural networks~(GNNs) are widely applied to learn graph node representations over graph-structured data. 
Current GNNs are generally built based on a neural message passing framework~\cite{gilmer2017neural} where one \prop operation and one \tran operation are performed in each layer.
The Graph Convolutional Network~(GCN) is one most representative GNN, which performs a graph convolution in the graph spectral domain per layer, with kernels approximated by a first-order Chebyshev polynomials~\cite{defferrard2016convolutional} of the normalized graph Laplacian matrix. 
GCNs have achieved high performance in the node classification task on various datasets~\cite{sen2008collective, hu2020ogb}.
In addition to GCNs, the GraphSage architecture~\cite{hamilton2017inductive} and Graph ATtention networks~(GATs)~\cite{velivckovic2017graph} are also popular GNNs. 
They are widely adopted for node classification,
achieving comparable performance with GCNs on many benchmark datasets~\cite{shchur2018pitfalls, hu2020ogb}.
GraphSage learn node representations by aggregating and transforming information from randomly sampled neighbors. 
GAT performs a learnable and flexible \prop by the attention mechanism~\cite{vaswani2017attention} in each layer.

\subsection{Performance degradation problem of deep GNNs}
It has been observed that existing GNN architectures tend to suffer performance degradation as their model depth increases~\cite{kipf2016semi, chen2019measuring}.
This problem is first observed in GCNs~\cite{kipf2016semi}, and later
Chen~\etal~\cite{chen2019measuring} find other popular GNN architectures such as GraphSage~\cite{hamilton2017inductive} and GAT~\cite{velivckovic2017graph} also suffer such performance degradation. 

Most studies on this problem are based on GCNs.
Existing works focus on how \props in GCNs affect node representations and cause performance degradation.
Li~\etal~\cite{li2018deeper} show that a \prop in GCN is essentially a Laplacian smoothing operation, and therefore \props push the node embeddings to be indistinguishable in deep GCNs, causing performance degradation, which is termed \textit{oversmoothing}. 
To reduce oversmoothing, Chen~\etal~\cite{chen2019measuring} introduce an additional loss to discourage similarity among distant nodes;
DropEdge~\cite{Rong2020DropEdge:} randomly removes edges from the graph during the training process;
PairNorm~\cite{zhao2019pairnorm} fixes the total pairwise feature distances across different layers.
In addition to oversmoothing, gradient vanishing is also identified by Li~\etal~\cite{li2019deepgcns} to be a reason for performance degradation, and is also widely believed to be caused by the smoothing effect of \props~\cite{li2019deepgcns, Rong2020DropEdge:, chen2020simple, li2020deepergcn}. 

Compared with \props, little attention has been paid to effects of \trans in this problem. 
Though Klicpera~\etal~\cite{klicpera2018predict} and Zhao \etal~\cite{zhao2019pairnorm} claim that \trans also make a reason for performance degradation, they do not investigate and justify it, and their focus is still on \props. 
To the best of our knowledge, \cite{oono2019asymptotic} is the only work that studies the role of \trans in the performance degradation problem.
However, their theoretical analysis is performed under the assumption that the input graph is sufficiently dense and the model depth goes to infinity, which is inapplicable to real world datasets or practical GCN models. 
In addition, their analysis is about how \trans and \props collectively lead to oversmoothing, rather than revealing whether and how \trans themselves contribute to performance degradation.
Unlike \cite{klicpera2018predict, zhao2019pairnorm, oono2019asymptotic}, our focus is placed on the role of \trans in performance degradation of GCNs.

In addition to the works investigating and addressing performance degradation in existing GNNs, there are also some works trying to design new GNN architectures that can naturally go deep without incurring severe performance drop.
A representative architectures among them is JKNet~\cite{xu2018representation}. 
However, as shown in \cite{Rong2020DropEdge:}, JKNet still suffers performance degradation when the model goes very deep (\eg 32- or 64-layer).
Concurrent to our work, GCNII~\cite{chen2020simple} and GEN~\cite{li2020deepergcn} show good performance even when they go very deep. 
GCNII addresses oversmoothing via initial residual connections and identity mappings, while GEN overcomes performance degradation with the help of several techniques including generalized aggregation functions, message normalization and layer normalization~\cite{ba2016layer}. 
This line of works, \ie JKNet, GCNII and GEN, are orthogonal and complementary to ours, as our work aims to better understand and resolve performance degradation based on the representative GCN architecture.
Furthermore, as we will show in Sec.~\ref{subsec:benefit-other-gnns}, applying our proposed variance-controlling technique to these architectures can also improve their performance.
   
\section{Understanding and resolving performance degradation of deep GCNs}
\label{sec:rethinking}

Neural networks usually perform better with increasing depth~\cite{he2016deep, telgarsky2016benefits}.
However, GCNs perform increasingly poorly with larger model depth.
Existing works mainly study how \props contribute to this problem, and pay little attention to the role of \trans, the other important operator that constitutes a GCN layer.

In this section, we study this problem from a new perspective. 
We start with ablative experiments to investigate the roles of \props and \trans in causing performance degradation. 
Based on the attained observations, we identify that the \varianceinflaming issue introduced by \trans is the critical contributing factor. 
Finally we develop a variance-controlling technique to alleviate this problem and improve performance of deep GCNs.

\subsection{Preliminaries}
\label{subsec:revisiting-preliminaries}

We first introduce the preliminaries on graph convolution network (GCN) models. 
Given an undirected graph $G$ with $n$ nodes, let the adjacency matrix and the degree matrix of $G$ be denoted as $A \in \{0,1\}^{n\times n}$ and $D=\mathrm{diag}(A\mathbf{1})$, 
where $\mathbf{1}$ is an $n$-dimensional all-ones column vector. 
An $L$-layer GCN model~\cite{kipf2016semi} is composed of $L$ cascaded feed-forward Graph Convolution~(GC) layers. 
Formally, the $l$-th GC layer can be represented as 
\begin{equation}
    \begin{aligned}
    H^{(l)} = \mathrm{ReLU}(\hat{A}H^{(l-1)}W^{(l)}),
    \end{aligned}
    \label{eqn:GC layer}
\end{equation}
where $H^{(l)}\in \mathbb{R}^{n\times d_l}$ and $W^{(l)}\in \mathbb{R}^{d_l \times d_{l+1}}$ denote the node feature matrix and the learnable weight matrix of this layer. 
The $i$-th row $(i\in \{1, \cdots, n\})$ of $H^{(l)}$, denoted as  $\mathbf{h}_i^{\top(l)}$, represents input embedding vector of node $i$ of the $l$-th layer. 
In this formula, $\hat{A}$ is the re-normalized adjacency matrix defined as $\hat{A}=\tilde{D}^{-\frac{1}{2}}\tilde{A}\tilde{D}^{-\frac{1}{2}}$, where $\tilde{A}=A+I$ and $\tilde{D} = \mathrm{diag}(\tilde{A}\mathbf{1})$. 
According to Eqn.~\eqref{eqn:GC layer}, one GCN layer performs the following two basic operations: the PROPagation operation~(\texttt{PROP}) and the TRANsformation operation~(\texttt{TRAN})~\cite{zhou2018graph, wu2019simplifying}:

\begin{equation}
    \begin{aligned}
    \bar{H}^{(l-1)} &= \hat{A}H^{(l-1)}\ \texttt{(PROP)}, \\
    H^{(l)} &= \mathrm{ReLU}(\bar{H}^{(l-1)}W^{(l)})\ 
    \texttt{(TRAN)}.
    \label{eqn:operations}
    \end{aligned}
\end{equation}
The first operation propagates and aggregates information from the 1-hop neighbors of each single node, while the latter transforms the aggregated embeddings via a linear transformation followed by a non-linear $\mathrm{ReLU}$~\cite{nair2010rectified} activation function.

A GCN model is built by stacking multiple GCN layers as above. 
Generally, with more GCN layers, the node feature information can be propagated to farther nodes~\cite{zhou2018graph,gilmer2017neural}. 
This is helpful for aggregating information from distant nodes, and hence improves the performance of GCNs~\cite{li2019deepgcns, zhao2019pairnorm, chen2020simple}.
However, some practical observations~\cite{li2018deeper, li2019deepgcns, chen2019measuring, Rong2020DropEdge:, zhao2019pairnorm}
are contradictory to this intuition\textemdash stacking more layers would incur severe performance drop for GCNs.

This is known as the performance degradation problem for deep GCNs. 
To address this problem, previous studies focus on \props, which are believed to cause the oversmoothing issue, \ie, node features in deep GCNs being pushed by \props to be indistinguishable from each other. 
Meanwhile, the role of \trans is largely overlooked in the previous studies. 
However, \trans are also critical for the performance of GCNs since they transform the aggregated information progressively and hence greatly influence the learned node representations.

\subsection{Transformation operations contribute significantly to performance degradation}
\label{subsec:trans-contribute-more}
\begin{figure*}[t!]
\centering
    \includegraphics[width=\linewidth]{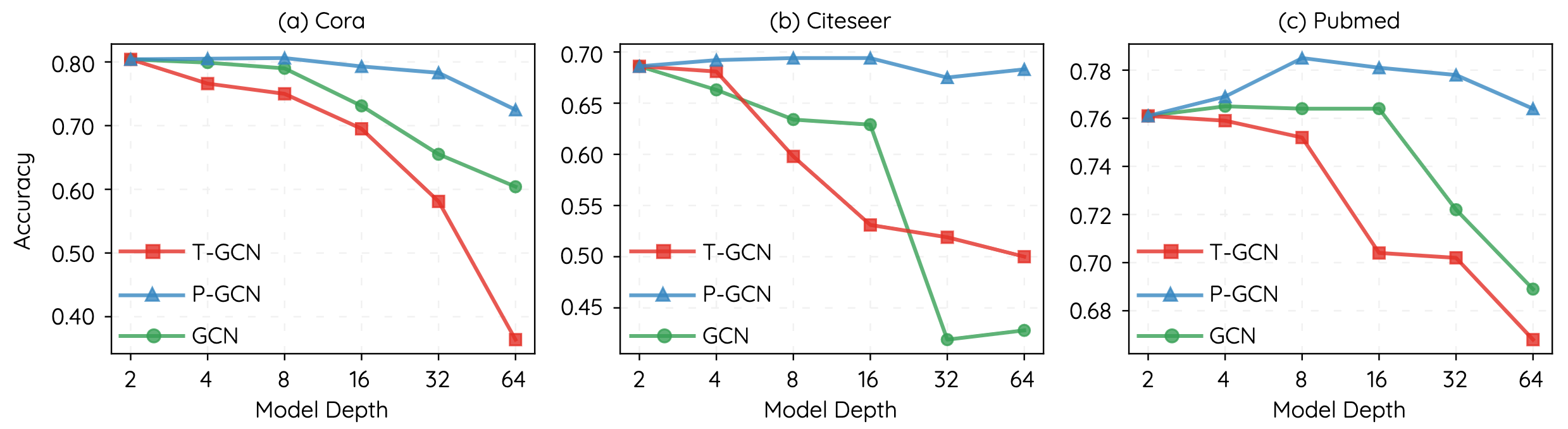}
    \caption{Performance degradation of GCNs, P-GCNs and T-GCNs. A T-GCN performs only \trans in hidden layers, while a P-GCN performs only \props in hidden layers. Due to limited space, we only show results on Cora, Citeseer and Pubmed.}
    \label{fig:trans-more}
\end{figure*}
To investigate the role of \trans in performance degradation, we need to exclude the influence of \props. 
To this end, we disentangle the two operations and build two variants of GCNs: 1)~T-GCNs that only perform \tran in each hidden layer; 2)~P-GCNs that only perform \prop in each hidden layer.

Formally, denote a GC layer as $\mathrm{GC}(\cdot)$ and a \texttt{TRAN} as $\mathrm{T(\cdot)}$, and let $X\in\mathbb{R}^{n\times d}$ be the input feature matrix.
Then, an $L$-layer T-GCN and P-GCN can be represented as
 \begin{equation}
    \label{eqn:stacking-operations-model}
    \begin{aligned}
    H^{(L)} &= \mathrm{GC}(\underbrace{\mathrm{T}\circ \cdots \circ \mathrm{T}}_{L-2}(\mathrm{GC}(X))) \quad \text{(T-GCN)}, \\
    H^{(L)} &= \mathrm{GC}(\hat{A}^{L-2}\mathrm{GC}(X)) \quad \text{(P-GCN)}.
    \end{aligned}
\end{equation}
Note the parameters of the two models in different layers are not shared. 

We train three models, \ie, vanilla GCN, P-GCN and T-GCN, of different depths on three benchmark datasets: Cora, Citeseer and Pubmed~\cite{sen2008collective}, and plot test accuracy in Fig.~\ref{fig:trans-more}.
Here the model depth varies from 2 to 64.
It can be seen that T-GCN suffers even more severe performance degradation than GCNs. \Eg, on Cora, the accuracy drops to 0.4 for 64-layer T-GCN, compared with the accuracy of 0.6 of 64-layer vanilla GCN; by contrast, although the performance of P-GCNs also drops when the the model goes very deep (\eg 64 layers), the accuracy only drops to 0.72. 
Such observations deviate from the conventional belief~\cite{li2018deeper, chen2019measuring} that more \props will hurt the model performance more severely.
Instead, we find that stacking more \trans introduces larger performance drop and \trans contribute more than \props in this study.

\subsection{Transformation operations cause variance inflammation}
\label{subsec:variance-inflaming}
\begin{figure*}[t!]
\centering
    \includegraphics[width=\linewidth]{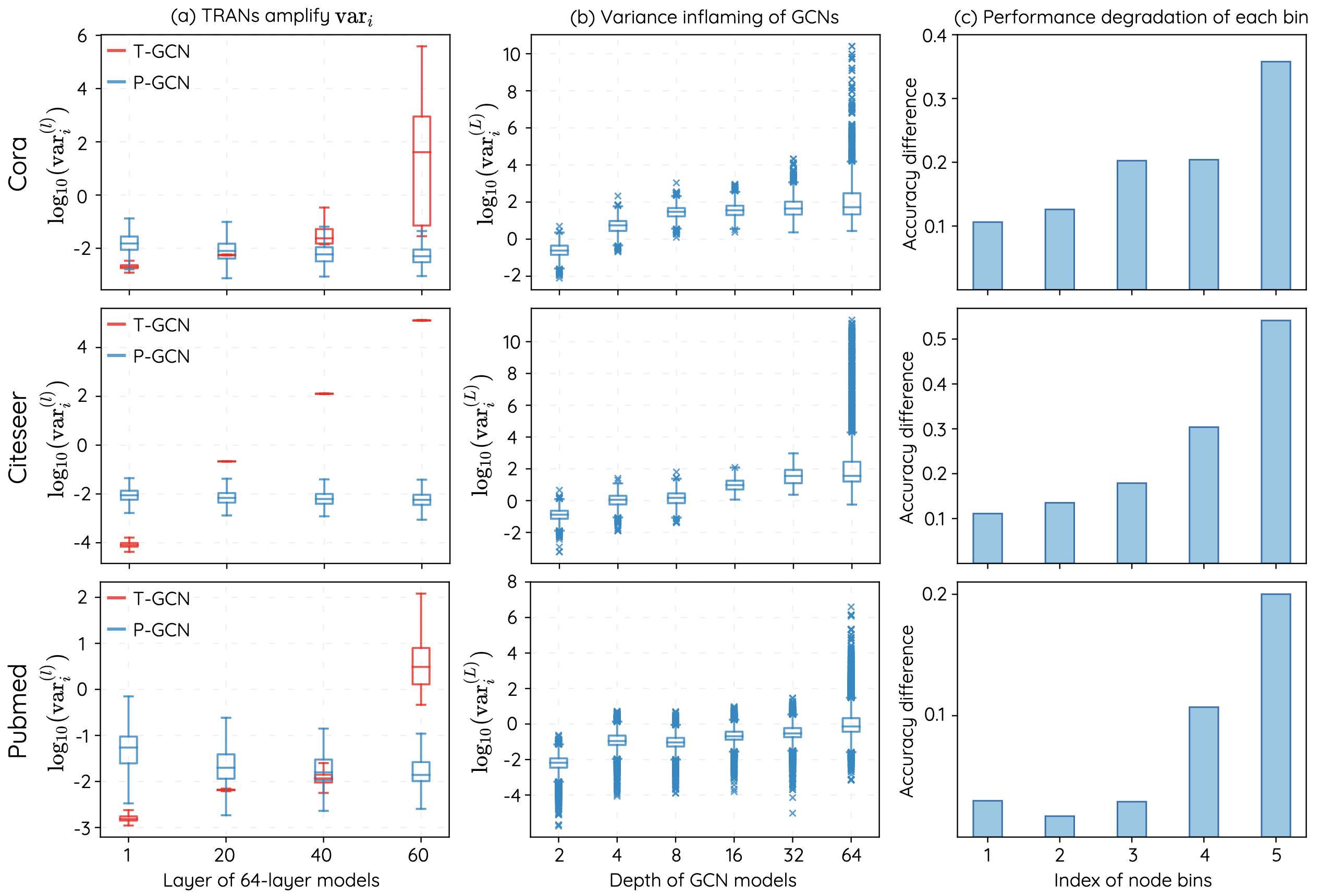}
    \caption{(a)~Node-wise feature variance, \ie, $\mathrm{var}_i^{(l)}$ of all nodes in different layers ($l=1,20,40,60$) of 64-layer T-GCN and P-GCN models. 
    Results are shown in log scale with a base of 10.
    Note that the two models are both 64-layer models. 
    Outlier points are not shown for better comparison of the two models. (b)~Node-wise feature variance of representations of the last layer, \ie, $\mathrm{var}_i^{(L)}$ (in log scale with a base of 10) for $L=2,4,68,16,32,64$ of all nodes. Note that models in this sub-figure are GCNs of different model depths.
    (c)~Performance difference between a 2-layer and a 64-layer model for each node bin. 
    From $S_1$ to $S_{5}$, $\mathrm{var}_i^{(L)}$ increases. Due to limited space, we only show results on Core, Citeseer and Pubmed.}
    \label{fig:variance_inflaming}
\end{figure*}
We then investigate why stacking more \trans would hurt the performance of deep models. 
We experimentally examine their effects on node representations, and find that \trans tend to amplify the node-wise feature variance. 
As a result, as a GCN model becomes deeper, it contains more \trans and hence its output node-wise feature variance becomes increasingly large in general. 
We refer to this phenomenon as \varianceinflaming. 
Here the node-wise feature variance refers to the variance of each node's features. 
Formally, the feature variance of node $i$ in the $l$-th layer is
\begin{equation}
    \begin{aligned}
    \mathrm{var}_i^{(l)} &= \frac{1}{d_l}\sum_{j=1}^{d_l}\left(h_{ij}^{(l)}-\mu_i^{(l)}\right)^2,
    \end{aligned}
    \label{eqn:node-wise-variance}
\end{equation}
where $h_{ij}$ is the the $j$-th feature of node $i$, $\mu_i^{(l)} = \frac{1}{d_l}\sum_{j=1}^{d_l}h_{ij}^{(l)}$ is the mean of the features, and $d_l$ denotes the feature dimension.

We plot how $\mathrm{var}_i^{(l)}$ changes with the layer index $l$ in a 64-layer T-GCN in Fig.~\ref{fig:variance_inflaming}~(a).
Results in 64-layer P-GCN are also included for comparison. 
As we can see, within a 64-layer T-GCN, the node-wise feature variance in general rises drastically (note that the y-axis is shown in log scale). 
By contrast, $\mathrm{var}_i^{(l)}$ in P-GCN does not show an increasing trend with larger $l$. This observation demonstrates that \trans tend to amplify node-wise feature variance.

We then plot the node-wise feature variance of the last layer of GCNs with different depths, \ie, $\mathrm{var}_i^{(L)}$ with different $L$, in Fig.~\ref{fig:variance_inflaming}~(b).
We can see \varianceinflaming from the drastic rise in $\mathrm{log(var}_i^{(L)})$ from $L=2$ to $L=64$, \ie such amplification of node feature variance leading to \varianceinflaming in GCNs.

Moreover, we find that nodes with large feature variance are difficult to classify. 
We observe this by sorting all the nodes in the graph based on their node-wise feature variance of the last layer, \ie, $\mathrm{var}_i^{(64)}$ of a 64-layer GCN, and partitioning the sorted nodes evenly into 5 bins: $S_1, \cdots, S_{5}$.
For each bin, we illustrate the performance degradation by calculating the accuracy difference between this 64-layer model and a 2-layer model. 
The results are summarized in Fig.~\ref{fig:variance_inflaming}~(c). 
We observe that the 64-layer model has relatively larger variance and performs significantly worse than the 2-layer one.

Motivated by these findings, we hypothesize that mitigating \varianceinflaming, \ie, preventing $\mathrm{var}_i$ from being too large, can help mitigate the performance degradation of deep GCNs.

\subsection{Techniques to mitigate variance inflammation}
\label{subsec:variance-controlling-techniques}
To reduce \varianceinflaming, we propose a technique that scales each node's feature vector by the $p$-th root of the standard deviation of its features, with $p \geqslant 1$.
As the operation is applied in a node-wise manner, we term it \emph{Node Normalization}~(\nodenorm), formally expressed as
\begin{equation}
    \label{eqn:nodenorm}
    \begin{aligned}
    \mathrm{NodeNorm}({\mathbf{h}_i}) = \frac{\mathbf{h}_i}{(\sigma_i)^{\frac{1}{p}}},
    \end{aligned}
\end{equation}
where $\sigma_i = \sqrt{\mathrm{var}_i}$ is the standard deviation of $\mathbf{h}_i$, and $\mathrm{var}_i$ is the variance defined in Eqn.~\eqref{eqn:node-wise-variance}.
Here we omit the layer index $l$ for clarity. 
In the following part, we use \nodenormp to denote \nodenorm with a specific $p$.

Let $\hat{\mathbf{h}}_i \triangleq \mathrm{NodeNorm}({\mathbf{h}_i})$ denote the normalized $\mathbf{h}_i$, and let $\hat {\mu}_i$, $\hat {\sigma}_i$ denote the mean and standard deviation of $\hat{\mathbf{h}}_i$.

We have
\begin{equation}
    \label{eqn: variance_normed}
    \begin{aligned}
    \hat{\sigma}_i &= \sqrt{\frac{1}{d}\sum_{j=1}^{d}\Big( \hat{h}_{ij} - \hat{\mu}_i\Big)^2} \\
    &=\sqrt{\frac{1}{d}\sum_{j=1}^{d}\Big( \frac{h_{ij}}{\sigma_i^{\frac{1}{p}}} - \frac{\mu_i}{\sigma_i^{\frac{1}{p}}}\Big)^2} \\
    &=\sigma_i^{(1-\frac{1}{p})}.
    \end{aligned}
\end{equation}

According to Eqn.~\eqref{eqn: variance_normed}, for large $\sigma_i$ ( $\sigma_i > 1$), $\sigma_i^{(1-\frac{1}{p})} \leqslant \sigma_i$ if $p \geqslant 1$. Therefore, given $p \geqslant 1$, we have: 1) The \nodenormp can reduce the variance of nodes with $\sigma_i > 1$, and hence alleviating \varianceinflaming; 2) A smaller $p$ controls \varianceinflaming more strictly. We can adjust the value of $p$ so that \nodenorm can collaborate well with GCNs in different scenarios. Specifically, $p=1$ yields the most strict \nodenorm, \ie, \nodenormone, which normalizes the variance for all nodes to be 1. 

Moreover, we note that the existing Layer Normalization~(LayerNorm)~\cite{ba2016layer} also performs a node-wise variance-scaling operation. 
We then also investigate whether (and why) LayerNorm is able to address performance degradation in GCNs.
Formally, the formulation of LayerNorm is:
\begin{equation}
    \begin{aligned}
    \mathrm{LayerNorm}({\mathbf{h}_i}) = \boldsymbol{\alpha} \odot \frac{\mathbf{h}_i - \mu_i}{\sigma_i} + \boldsymbol{\beta},
    \end{aligned}
    \label{eqn:layernorm}
\end{equation}
where $\odot$ denotes an element-wise multiplication, and $\boldsymbol{\alpha}$, $\boldsymbol{\beta}$ are learnable parameters. We can see that a LayerNorm consists of three operations: variance-scaling, mean-subtraction, and feature-wise linear transformation ($\boldsymbol{\alpha}$, $\boldsymbol{\beta}$ are the slopes and biases). In particular, the variance-scaling in LayerNorm is essentially our \nodenormone. 

One may argue that LayerNorm does not naturally mitigates \varianceinflaming, due to the linear transformations.
However, we empirically find that LayerNorm operations in deep GCNs are trained to effectively reduce \varianceinflaming (see Sec.~\ref{subsec: investigate-layernorm}).

\section{Experiments}
\label{sec:experiments}
As observed in existing works~\cite{zhao2019pairnorm, Rong2020DropEdge:}, deep GCNs do not show significant advantage over shallow ones on benchmark settings 
(\eg, 20-label-per-class setting on Cora). This might be because shallow models are already sufficient to learn good node representations in these settings, as discussed in~\cite{zhao2019pairnorm}. Therefore, to demonstrate the benefit of mitigating \varianceinflaming, we first evaluate \nodenorm in three exemplar cases where shallow models are insufficient in Sec.~\ref{subsec:allow-deep-perform-better} (\ie, deep models are needed). We then validate whether the proposed \nodenorm can help alleviating performance degradation of GCNs on benchmark settings in Sec.~\ref{subsec:address-performance-degradation}. Next, in Sec.~\ref{subsec:benefit-other-gnns}, we apply our proposed \nodenorm to other GNN architectures to study their effects upon the performance of different GNNs. Finally, we investigate the effectiveness of LayerNorm in alleviating performance degradation.

\begin{table*}[t!]
    \caption{Classification accuracy in cases where deeper models are needed. The first row is result of the original GCNs. Following rows correspond to GCNs with equipped different techniques. The number in parenthesis corresponds to depth of the model that achieves the performance.}
    \label{tab:gcn-deeper-needed}
    \centering
    \resizebox{1.0\linewidth}{!}{
    \begin{small}
    \begin{tabular}{|c|c|c|c|c|c|c|c|c|}
    \hline
          \multirow{2}{*}{} & \multicolumn{3}{c|}{Citation Networks w/ missing features} & \multicolumn{3}{c|}{Citation Networks w/ low label rate} & \multicolumn{2}{c|}{Networks w/ large diameter} \\
          \cline{2-9}
          & Cora & Citeseer & Pubmed & Cora & Citeseer & Pubmed & USelect-12 & USelect-16\\
          \hline
          GCN & 0.7034{\scriptsize \gray{$\pm$0.0235}}  (8) &  0.4494{\scriptsize \gray{$\pm$0.0227}}   (8) &  0.4652{\scriptsize \gray{$\pm$0.0677}}  (16) & 0.6319{\scriptsize \gray{$\pm$0.0982}}   (4)  &  0.5277{\scriptsize \gray{$\pm$0.0695}}   (2)   &  0.6491{\scriptsize \gray{$\pm$0.0675}}   (4) &   0.8296{\scriptsize \gray{$\pm$0.0147}}   (2) &0.8840{\scriptsize \gray{$\pm$0.0094}}  (2)\\
          \hline
           +DropEdge &  0.7335{\scriptsize \gray{$\pm$0.0182}}  (16) &   0.4811{\scriptsize \gray{$\pm$0.0225}}  (16)  &   0.4292{\scriptsize \gray{$\pm$0.0331}}  (32)  &   0.6193{\scriptsize \gray{$\pm$0.0496}}  (4)  &   0.4710{\scriptsize \gray{$\pm$0.0679}}  (8)  &   0.6557{\scriptsize \gray{$\pm$0.0717}}  (4)  &   0.8589{\scriptsize \gray{$\pm$0.0082}}  (16)  &0.8921{\scriptsize \gray{$\pm$0.0037}}  (4) \\
           +PairNorm &  0.6947{\scriptsize \gray{$\pm$0.0230}}  (64) &   0.4475{\scriptsize \gray{$\pm$0.0201}}  (32)  &   \textbf{0.6683}{\scriptsize \gray{$\pm$0.0387}}  (32)  &   0.6168{\scriptsize \gray{$\pm$0.0624}}  (16)  &   0.4924{\scriptsize \gray{$\pm$0.0503}}  (8)  &   0.6445{\scriptsize \gray{$\pm$0.0663}}  (32)  &   0.8665{\scriptsize \gray{$\pm$0.0113}}  (32)  &0.8983{\scriptsize \gray{$\pm$0.0098}}  (32) \\
          \hline
          +\nodenormone &   0.7207{\scriptsize \gray{$\pm$0.0122}}  (64) &   0.4861{\scriptsize \gray{$\pm$0.0224}}  (32)  &   0.5751{\scriptsize \gray{$\pm$0.0504}}  (16)  &   0.6420{\scriptsize \gray{$\pm$0.0301}}  (16) &   0.5516{\scriptsize \gray{$\pm$0.0702}}  (16)  &   0.6813{\scriptsize \gray{$\pm$0.0350}}  (64)  &   0.8677{\scriptsize \gray{$\pm$0.0099}}  (32)  &0.9017{\scriptsize \gray{$\pm$0.0063}}  (32) \\
          +\nodenormtwo &  0.7361{\scriptsize \gray{$\pm$0.0180}}  (16) &   \textbf{0.4957}{\scriptsize \gray{$\pm$0.0199}}  (32)  &   0.6106{\scriptsize \gray{$\pm$0.0387}}  (16)  &   \textbf{0.6605}{\scriptsize \gray{$\pm$0.0420}}  (16)  &   0.5551{\scriptsize \gray{$\pm$0.0685}}  (4)  &   \textbf{0.6908}{\scriptsize \gray{$\pm$0.0432}}   (32)  &   \textbf{0.8700}{\scriptsize \gray{$\pm$0.0138}}  (32)  & \textbf{0.9028}{\scriptsize \gray{$\pm$0.0128}}  (32) \\
          +\nodenormthree &  \textbf{0.7395}{\scriptsize \gray{$\pm$0.0312}}  (16) &   0.4683{\scriptsize \gray{$\pm$0.0311}}  (8)  &   0.4713{\scriptsize \gray{$\pm$0.0626}}  (32)  &   0.6580{\scriptsize \gray{$\pm$0.0544}}  (8)  &   \textbf{0.5618}{\scriptsize \gray{$\pm$0.0715}}  (4)  &   0.6694{\scriptsize \gray{$\pm$0.0619}}  (8)  &   0.8697{\scriptsize \gray{$\pm$0.0138}}  (32)  &0.9026{\scriptsize \gray{$\pm$0.0093}}  (32) \\

          \hline 
    \end{tabular}
    \end{small}
    }
\end{table*}

\subsection{Evaluating effects of proposed \nodenorm in cases requiring deep models}
\label{subsec:allow-deep-perform-better}
In this subsection, we evaluate \nodenorm in three exemplar cases where deep models are needed to learn good node representations.

\subsubsection{Experiment settings}
\label{subsubsec:deep-better-setting}
We first introduce the exemplar cases:\\
\begin{enumerate}
    \item \textbf{Citation networks with missing features}. In~\cite{zhao2019pairnorm}, when some  input node features are missing in citation graphs, deep models achieve better performance than shallow ones.
    Here shallow models are not sufficient to learn good node representations because nodes would benefit from a larger neighbourhood to recover effective feature representation.
    \item \textbf{Citation graphs with low label rate}. 
    Deeper models achieve better performance than shallow ones on Cora, Citeseer and Pubmed at low training label rate. 
    As explained in~\cite{sun2019multi}, when training label rate is low, more layers would be needed to reach the supervision information far away.
    \item  \textbf{US election datasets}.
    USelect-12 and USelect-16 datasets~\cite{jia2020residual} are geographical graphs induced from statistics of United States~(US) election of year 2012 and year 2016.
    Nodes represent US counties, and edges connect nodes whose corresponding counties are geographically bordering.
    Node features are demographic statistics such as income, education, population. 
    The graph structures of the two datasets are exactly the same. 
    Deep model significantly outperform shallow ones on them, possibly because their graph has a diameter of 69, which is notably larger than that of the commonly used datasets (\eg about 20).
    In addition, the average shortest path length between node pairs in the two datasets is around 26, over 4 times larger than that of citation networks.
    Given such a large graph diameter and a long average distance among nodes, deep models would be desired to fully propagate information among nodes and learn good representations.
\end{enumerate}

For the missing-feature case, we follow \cite{zhao2019pairnorm} to run experiments on Cora, Citeseer and Pubmed with 100\% of missing features. We follow widely adopted 20-label-per-class setting~\cite{kipf2016semi, zhao2019pairnorm}: 20 labeled training nodes per class, 500 validation nodes, 1,000 test nodes. For the case of low label rate citation graphs, we run experiments on Cora, Citeseer and Pubmed with the 2-label-per-class setting (label rates for three datasets are 0.52\%, 0.36\%, and 0.03\%). 
The sizes of validation and test sets are 500 and 1,000, as in the commonly adopted 20-label-per-class setting.
For the US election datasets, we randomly split the nodes into train/val/test by 60\%/20\%/20\%~\cite{jia2020residual}, and follow \cite{huang2020combining} to conduct a binary node classification task. 

For baseline methods, we choose PairNorm~\cite{zhao2019pairnorm} and DropEdge~\cite{Rong2020DropEdge:}, which are the best performing generic (\ie, plug-in) methods to address performance degradation of GCNs and other GNN architectures. For DropEdge, results are obtained with the PyTorch Geometric library~\cite{Fey/Lenssen/2019} implementation of DropEdge; for PairNorm, results are obtained using their official implementation. Here we do not compare works designing new GNN architectures that suffer little performance degradation~\cite{li2020deepergcn, chen2020simple}, since they are orthogonal to our design. Instead, we discuss these methods in Sec.~\ref{subsec:benefit-other-gnns}.

We run experiments with GCNs of \{2,4,8,16,32,64\} layers. To make the results more reliable, we run each experiment with 10 different random splits of the dataset and report the mean and standard deviation, as suggested by \cite{shchur2018pitfalls}. We add residual connections~\cite{he2016deep, li2018deeper} in each layer to avoid training difficulty caused by gradient vanishing. We use 64 hidden dimension for all methods.
For \nodenorm, we run experiments with $p=1,2,3$.

\subsubsection{Results}
\label{subsubsec:deeper-needed-case-results}
Tab.~\ref{tab:gcn-deeper-needed} summarizes the best average accuracy and corresponding model depth of the experiments above.
In general, with the help of \nodenorm, deeper models achieve much better performance than shallow ones. Specifically, the best performance of \nodenorm-augmented GCNs is achieved by deep models (\eg 64-layer), and is higher than the best performance of vanilla GCNs, which is generally obtained by shallow models (\eg 2-layer). 

Though both vanilla GCNs and \nodenorm-augmented GCNs are capable of aggregating information from distant nodes when their model depth increases~\cite{gilmer2017neural, hamilton2017inductive}, vanilla GCNs do not perform better as they become deeper. 
This is because \varianceinflaming (along with other factors like oversmoothing) offsets the benefit of aggregating distant node information and even leads to performance degradation. By contrast, deep GCNs augmented with \nodenorm successfully outperform their shallow counterparts because \nodenorm resolves \varianceinflaming.

Tab.~\ref{tab:gcn-deeper-needed} also show that the baseline methods, \ie, DropEdge and PairNorm, are able to alleviate performance degradation of deep models to some extent, but they are inferior to our proposed variance \nodenorm.

Among \nodenorm with different $p$, \nodenormtwo has the best performance in 5 out of 8 scenarios. \nodenormthree also stands out in two cases. In addition, \nodenormone is also among the top-3 best-performing techniques in most scenarios. Furthermore, we show in the next section that \nodenormone is the most effective one when the model goes very deep (\eg, 64-layers).


Another interesting observation is that, with varying $p$, the optimal model depth can be different. In general, smaller $p$ corresponds to a larger optimal model depth. One possible explanation is that shallower models suffer less \varianceinflaming, thus requiring less strict variance controlling to achieve better performance. 

\begin{figure*}[t!]
\centering
    \includegraphics[width=\linewidth]{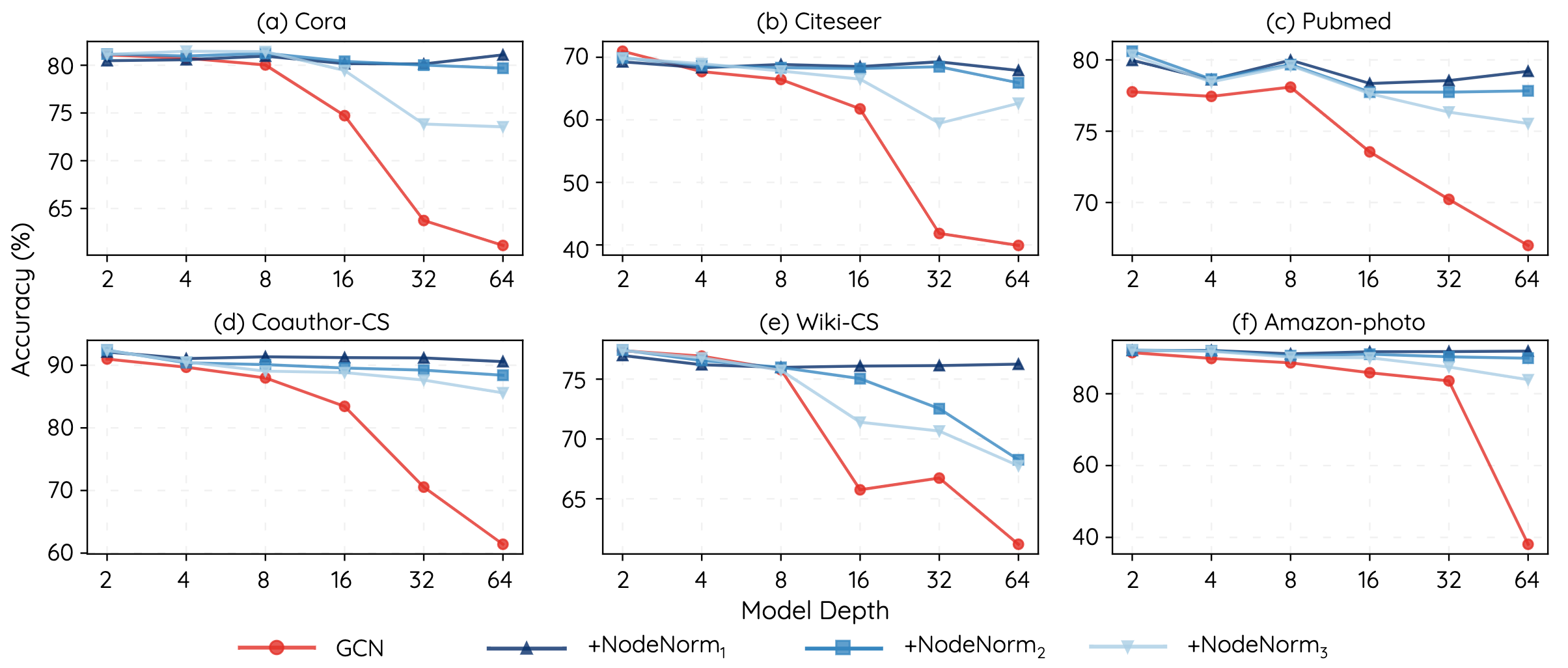}
    \caption{Classification accuracy \wrt model depth. It can be seen all variance-controlling techniques are able to mitigate performance degradation.  Our \nodenormone enables deep models to compete with shallow ones by strictly controlling node-wise variance, \ie, ensuring all $\mathrm{var}_i$ to be 1}
 
    \label{fig:address-performance-degradation}
\end{figure*}
\subsection{Evaluating effects of proposed \nodenorm on benchmark datasets}
\label{subsec:address-performance-degradation}
To further investigate whether \nodenorm can help resolve performance degradation of deep GCNs, we evaluate the proposed \nodenorm on 6 commonly used benchmark datasets.

We first run experiments on 6 node classification datasets of various graph types: three benchmark citation datasets~\cite{sen2008collective} Cora, Citeseer and Pubmed, a co-authorship dataset Coauthor-CS~\cite{shchur2018pitfalls}, a web-page dataset Wiki-CS~\cite{mernyei2020wiki}, and a product co-purchasing dataset Amazon-photo~\cite{shchur2018pitfalls}. For the three citation networks Cora, Citeseer and Pubmed, we follow the widely adopted 20-label-per-class setting~\cite{kipf2016semi, velivckovic2017graph, xu2018representation, chen2019measuring}.
For Coauthor-CS, we also run experiments with 10 splits that are randomly generated based on splitting rules in~\cite{verma2019graphmix}. 
For Wiki-CS, we use the 20 splits provided in \cite{mernyei2020wiki}. 
For Amazon-photo, following~\cite{shchur2018pitfalls} we adopt the same split setting as the three citation networks.
The evaluation metric for these datasets is classification accuracy. We run experiment with GCNs with \{2, 4, 8, 16, 32, 64\} layers and use $p=1,2,3$ for \nodenormp.
\begin{table}[t!]
    \caption{Classification accuracy of \nodenormone, PairNorm and Dropedge on a widely used split of Cora, Citeseer and Pubmed. }
    \label{tab:compare-with-sota-exisiting}
    \centering
    \resizebox{0.75\linewidth}{!}{
    \begin{scriptsize}
    \begin{tabular}{|c|c|c|c|c|}
    \hline
           \multirow{2}{*}{Dataset} & \multirow{2}{*}{Method} & \multicolumn{3}{c|}{Model depth} \\
           \cline{3-5}
           & & 2 & 32 & 64 \\ 
          \hline
          \multirow{3}{*}{Cora} & \nodenormone & \textbf{0.830} & \textbf{0.829} & \textbf{0.837} \\
          \cline{2-5}
          & PairNorm & 0.783 & 0.759 & 0.778 \\
          \cline{2-5}
          & DropEdge & 0.828 & 0.811 & 0.789 \\
          \hline
          \multirow{3}{*}{Citeseer} & \nodenormone & \textbf{0.729} & \textbf{0.724} & \textbf{0.731} \\
          \cline{2-5}
          & PairNorm & 0.648 & 0.615 & 0.614 \\
          \cline{2-5}
          & DropEdge & 0.723 & 0.700 & 0.651\\
          \hline
          \multirow{3}{*}{Pubmed} & \nodenormone & \textbf{0.807} & \textbf{0.808} & \textbf{0.804} \\
          \cline{2-5}
          & PairNorm & 0.756 & 0.768 & 0.737 \\
          \cline{2-5}
          & DropEdge & 0.796 & 0.782 & 0.769 \\
          \hline
    \end{tabular}
    \end{scriptsize}
    }
\end{table}

As shown in Fig.~\ref{fig:address-performance-degradation}, compared with vanilla GCNs, models with \nodenorm suffer much less accuracy drop when they get deep. In particular, \nodenormone enables very deep models (\eg, 64-layer) to achieve comparable performance with shallow ones (\eg, 2-layer). The results well demonstrate that mitigating \varianceinflaming indeed helps address performance degradation. 

This is further justified by comparing \nodenormp with varying $p$.
For deep models (\eg 64-layers) that suffer severe \varianceinflaming, the model performance increases as $p$ decreases from 3 to 1. 
The comparison shows that techniques that control \varianceinflaming more strictly (with a smaller $p$) can address performance degradation more effectively for very deep models. 

Furthermore, we compare our \nodenormone, the most effective \nodenorm for very deep models, with the baseline DropEdge~\cite{Rong2020DropEdge:} and PairNorm~\cite{zhao2019pairnorm} methods on benchmark datasets. For fair comparison, we follow \cite{Rong2020DropEdge:, zhao2019pairnorm} to run experiments on Cora, Citeseer and Pubmed in
the 20-label-per-class semi-supervised classification setting,
with a widely used standard split \cite{kipf2016semi, velivckovic2017graph}. Results for DropEdge are from their GitHub repository. For PairNorm, 2-layer results are reported in their paper, and we reproduce other results using their reported settings.

Tab~\ref{tab:compare-with-sota-exisiting} summarizes the results. We can see though Dropedge and PairNorm alleviate performance degradation to some extent, deep models with these methods still
underperform their shallow counterparts. In comparison,
our \nodenormone successfully addresses the performance
degradation, enabling GCNs of 32 or 64 layers to achieve
comparable results with 2-layer GCNs.

Note that in the above results, deep GCNs do not show significant advantage over shallow ones. It is because in benchmark settings, shallow models might be sufficient to learn good representations (discussed in \cite{zhao2019pairnorm}). Nevertheless, our \nodenorm successfully enables deep models to match the performance of shallow ones.
\begin{figure*}[t!]
\centering
    \includegraphics[width=0.90\linewidth]{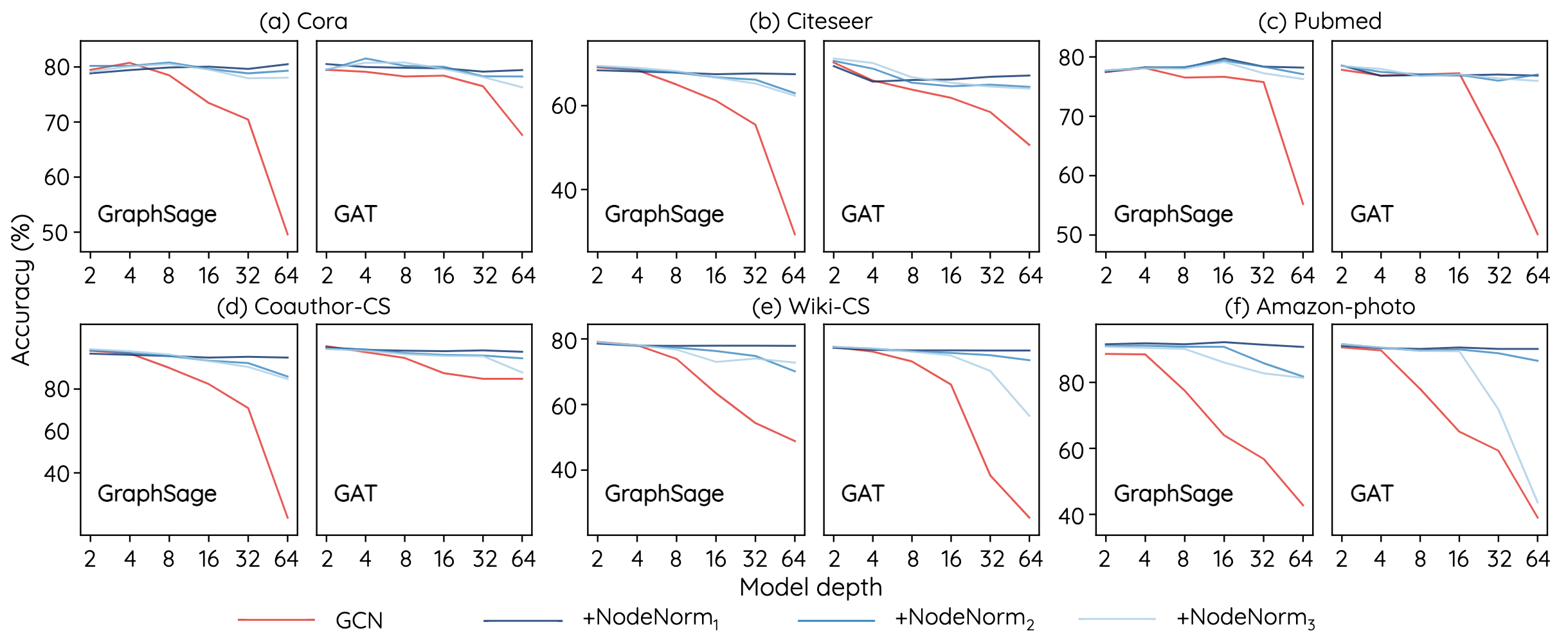}
    \caption{Classification accuracy \wrt model depth of GAT and GraphSage.}
    \label{fig:generalize-to-other}
\end{figure*}

\subsection{Applying variance-controlling to other GNN architectures}
\label{subsec:benefit-other-gnns}
\begin{table}[t!]
    \caption{Best AUC-ROC and corresponding model depth on Obgn-proteins of GENs with and without \nodenormone.
    }
    \label{tab:improving-gen}
    \centering
\resizebox{0.5\linewidth}{!}{
\begin{tabular}{|c|c|}
\hline
Method & AUC-ROC \\
\hline
GEN & 0.7936{\scriptsize \gray{$\pm$0.0086}}  (64)  \\
\hline

+\nodenormone & \textbf{0.8226}{\scriptsize \gray{$\pm$0.0093}}  (64)  \\
\hline
\end{tabular}}
\end{table}

We apply \nodenorm to other GNN architectures to study whether variance-controlling can further improve their performance. 
Below we first experiment on GNNs that also suffer performance degradation when the model gets deep~\cite{hamilton2017inductive, velivckovic2017graph}, and then on those that can go deep without performance drop~\cite{chen2020simple, li2020deepergcn}. 

We first apply the variance-controlling techniques (\ie \nodenorm) to other popular GNN architectures that also suffer from performance degradation, including GAT~\cite{velivckovic2017graph} and GraphSage~\cite{hamilton2017inductive}, which achieve high performance in the node classification task. 
We conduct experiments using the same setting as in Sec.~\ref{subsec:address-performance-degradation}.
As shown in Fig.~\ref{fig:generalize-to-other}, the variance-controlling techniques generalize well to these architectures, mitigating their performance degradation effectively.
In particular, \nodenorm enables deep GAT or GraphSage models to compete with their shallow counterparts. 
The results well demonstrate that controlling variance also addresses performance degradation for other GNN architectures in addition to GCNs.

\begin{table}[t!]
\caption{Best accuracy and corresponding model depth on web-pages datasets of GCNIIs with and without \nodenormone.} 
\label{tab:improving-gcnii}
\centering
\resizebox{1.0\linewidth}{!}{
 \begin{tabular}{|c|c|c|c|}
\hline
\multirow{2}{*}{Method} & \multicolumn{3}{c|}{Dataset} \\
\cline{2-4}
& Cornell & Texas & Wisconsin\\
\hline
GCNII & 0.7496{\scriptsize \gray{$\pm$0.0514}}  (16) & 0.6946{\scriptsize \gray{$\pm$0.0793}} (32) & 0.7412{\scriptsize \gray{$\pm$0.0526}} (16) \\
\hline
+\nodenormone & \textbf{0.8054}{\scriptsize \gray{$\pm$0.0692}}  (16) & \textbf{0.7892}{\scriptsize \gray{$\pm$0.0637}} (32) & \textbf{0.8314}{\scriptsize \gray{$\pm$0.0542}} (16) \\
\hline
\end{tabular}}
\end{table}

There are also works obtaining deep GNNs by designing \textit{new} GNN architectures that can naturally go deep without suffering severe performance degradation, which are orthogonal to our design in this work. 
We also include them in experiments to further show the effectiveness of our proposed variance controlling techniques.
In particular, we apply \nodenorm to GEN~\cite{li2020deepergcn} and GCNII~\cite{chen2020simple}, which are among the best new GNN architectures that do not suffer declined performance as the model grows deep.
We run experiments with the two models of \{2, 4, 8, 16, 32, 64\} layers.
For GEN, we follow \cite{li2020deepergcn} to run experiments on Ogbn-proteins~\cite{hu2020ogb}, each with 5 random seeds, since the dataset split in~\cite{hu2020ogb} is practically meaningful and random split may not make sense anymore. We use the PyTorch Geometric library~\cite{Fey/Lenssen/2019} implementation of GENs, which is provided by \cite{li2020deepergcn}. 
For GCNII, we follow \cite{chen2020simple} to conduct experiments on web-page networks~\cite{Pei2020Geom-GCN:}: Cornell, Texas and Wisconsin, where each experiment is run with 10 different dataset splits. 
In the web-page networks, nodes and edges represent web pages and hyperlinks respectively. 
We use their released code for implementation.
For both architectures, we follow experiment settings in their GitHub repositories. We do not tune hyperparameters or use training tricks like one-hot-node-encoding in \cite{li2020deepergcn}. We take \nodenormone as an example of \nodenorm as it is simple yet effective in addressing performance degradation for very deep models.


As shown in Tab.~\ref{tab:improving-gen} and Tab.~\ref{tab:improving-gcnii}, our proposed \nodenormone improve the performance of GEN and GCNII. Notably, \nodenormone improves the performance of GCNII by a margin of 10\% on Texas and Wisconsin datasets.
We emphasize again that GEN and GCNII are orthogonal to our work\textemdash we focus on understanding and addressing performance degradation of GCNs (and other popular GNNs), and claim that reducing \varianceinflaming helps mitigate this issue, while their contributions are the proposed architectures, \ie, GEN or GCNII. 

\subsection{Investigating the effectiveness of $\mathrm{LayerNorm}$}
\label{subsec: investigate-layernorm}
As mentioned in Sec.~\ref{subsec:variance-controlling-techniques}, LayerNorm~\cite{ba2016layer} also performs a variance-scaling operation. Inspired by this, we investigate whether and why LayerNorm also helps address performance degradation in GCNs.

We experiment with LayerNorm in the three exemplar cases where deep GCNs are desired (see Sec.~\ref{subsec:allow-deep-perform-better}), using the same setting as in Sec.~\ref{subsec:allow-deep-perform-better}. As shown in Tab.~\ref{tab:compare-layernorm}, LayerNorm also helps deeper models to achieve better performance than shallow ones. 

Meanwhile, we note that, our proposed \nodenorm outperform LayerNorm in these experiments: \nodenormone, \nodenormtwo and \nodenormthree outperform LayerNorm in 5, 7 and 5 out of 8 settings respectively. This demonstrates that \nodenorm is more effective than LayerNorm in resolving performance degradation. 
One possible reason might be that less strict variance controlling is desired in these settings. Therefore, techniques that control node-wise more softly (\eg, \nodenormtwo) would outperform those strictly controlling the variance, such as \nodenormone and LayerNorm (note that LayerNorm performs the variance-scaling operation that is equivalent to \nodenormone).
Another possible reason is that LayerNorm has learnable parameters $\mathbf{\alpha}$ and $\mathbf{\beta}$ (see Eqn.~\eqref{eqn:layernorm}), which increases the degree of overfitting and hence slightly hurts the performance.

\begin{table*}[t!]
    \caption{Performance comparison of LayerNorm and \nodenorm in cases where deeper models are needed. The first row shows the results of the original GCNs. Following rows correspond to GCNs with different techniques equipped. The number in parenthesis corresponds to depth of the model that achieves the performance.}
    \label{tab:compare-layernorm}
    \centering
    \resizebox{1.0\linewidth}{!}{
    \begin{small}
    \begin{tabular}{|c|c|c|c|c|c|c|c|c|}
    \hline
          \multirow{2}{*}{} & \multicolumn{3}{c|}{Citation Networks w/ missing features} & \multicolumn{3}{c|}{Citation Networks w/ low label rate} & \multicolumn{2}{c|}{Networks w/ large diameter} \\
          \cline{2-9}
          & Cora & Citeseer & Pubmed & Cora & Citeseer & Pubmed & USelect-12 & USelect-16\\
          \hline
          GCN & 0.7034{\scriptsize \gray{$\pm$0.0235}}  (8) &  0.4494{\scriptsize \gray{$\pm$0.0227}}   (8) &  0.4652{\scriptsize \gray{$\pm$0.0677}}  (16) & 0.6319{\scriptsize \gray{$\pm$0.0982}}   (4)  &  0.5277{\scriptsize \gray{$\pm$0.0695}}   (2)   &  0.6491{\scriptsize \gray{$\pm$0.0675}}   (4) &   0.8296{\scriptsize \gray{$\pm$0.0147}}   (2) &0.8840{\scriptsize \gray{$\pm$0.0094}}  (2)\\

          +LayerNorm &  0.7160{\scriptsize \gray{$\pm$0.0199}}  (64) &   0.4860{\scriptsize \gray{$\pm$0.0180}}  (32)  &   0.5450{\scriptsize \gray{$\pm$0.0577}}  (4)  &   0.6348{\scriptsize \gray{$\pm$0.0385}}  (16)  &   0.5569{\scriptsize \gray{$\pm$0.0747}}  (16)  &   0.6820{\scriptsize \gray{$\pm$0.0464}}  (32)  &   0.8657{\scriptsize \gray{$\pm$0.0104}}  (32)  &0.8983{\scriptsize \gray{$\pm$0.0113}}  (32) \\
          \hline

          +\nodenormone &   0.7207{\scriptsize \gray{$\pm$0.0122}}  (64) &   0.4861{\scriptsize \gray{$\pm$0.0224}}  (32)  &   0.5751{\scriptsize \gray{$\pm$0.0504}}  (16)  &   0.6420{\scriptsize \gray{$\pm$0.0301}}  (16) &   0.5516{\scriptsize \gray{$\pm$0.0702}}  (16)  &   0.6813{\scriptsize \gray{$\pm$0.0350}}  (64)  &   0.8677{\scriptsize \gray{$\pm$0.0099}}  (32)  &0.9017{\scriptsize \gray{$\pm$0.0063}}  (32) \\
          +\nodenormtwo &  0.7361{\scriptsize \gray{$\pm$0.0180}}  (16) &   \textbf{0.4957}{\scriptsize \gray{$\pm$0.0199}}  (32)  &   \textbf{0.6106}{\scriptsize \gray{$\pm$0.0387}}  (16)  &   \textbf{0.6605}{\scriptsize \gray{$\pm$0.0420}}  (16)  &   0.5551{\scriptsize \gray{$\pm$0.0685}}  (4)  &   \textbf{0.6908}{\scriptsize \gray{$\pm$0.0432}}   (32)  &   \textbf{0.8700}{\scriptsize \gray{$\pm$0.0138}}  (32)  & \textbf{0.9028}{\scriptsize \gray{$\pm$0.0128}}  (32) \\
          +\nodenormthree &  \textbf{0.7395}{\scriptsize \gray{$\pm$0.0312}}  (16) &   0.4683{\scriptsize \gray{$\pm$0.0311}}  (8)  &   0.4713{\scriptsize \gray{$\pm$0.0626}}  (32)  &   0.6580{\scriptsize \gray{$\pm$0.0544}}  (8)  &   \textbf{0.5618}{\scriptsize \gray{$\pm$0.0715}}  (4)  &   0.6694{\scriptsize \gray{$\pm$0.0619}}  (8)  &   0.8697{\scriptsize \gray{$\pm$0.0138}}  (32)  &0.9026{\scriptsize \gray{$\pm$0.0093}}  (32) \\

          \hline 
    \end{tabular}
    \end{small}
    }
\end{table*}

However, since LayerNorm performs three operations: variance-scaling, mean-subtraction and linear transformation (see Eqn.~\eqref{eqn:layernorm}), more investigation is needed to demonstrate that variance-scaling is the key to the effectiveness of LayerNorm in resolving performance degradation of deep GCNs. 
We then ablatively study the effects of the three operations on addressing performance degradation.
Specifically, we study two variants of LayerNorm: 
\begin{equation}
    \begin{aligned}
    \mathrm{LayerNorm}^*({\mathbf{h}_i}) = \frac{\mathbf{h}_i - \mu_i}{\sigma_i},
    \end{aligned}
    \label{eqn:layernorm*}
\end{equation}
\begin{equation}
    \begin{aligned}
    \mathrm{LayerNorm\text{-}MS}({\mathbf{h}_i}) = \mathbf{h}_i - \mu_i.
    \end{aligned}
    \label{eqn:layernorm-ms}
\end{equation}
LayerNorm* does not include linear transformations, while the LayerNorm-MS variant performs only Mean-Subtraction (MS). 

\begin{figure}
    \centering
    \includegraphics[width=\linewidth]{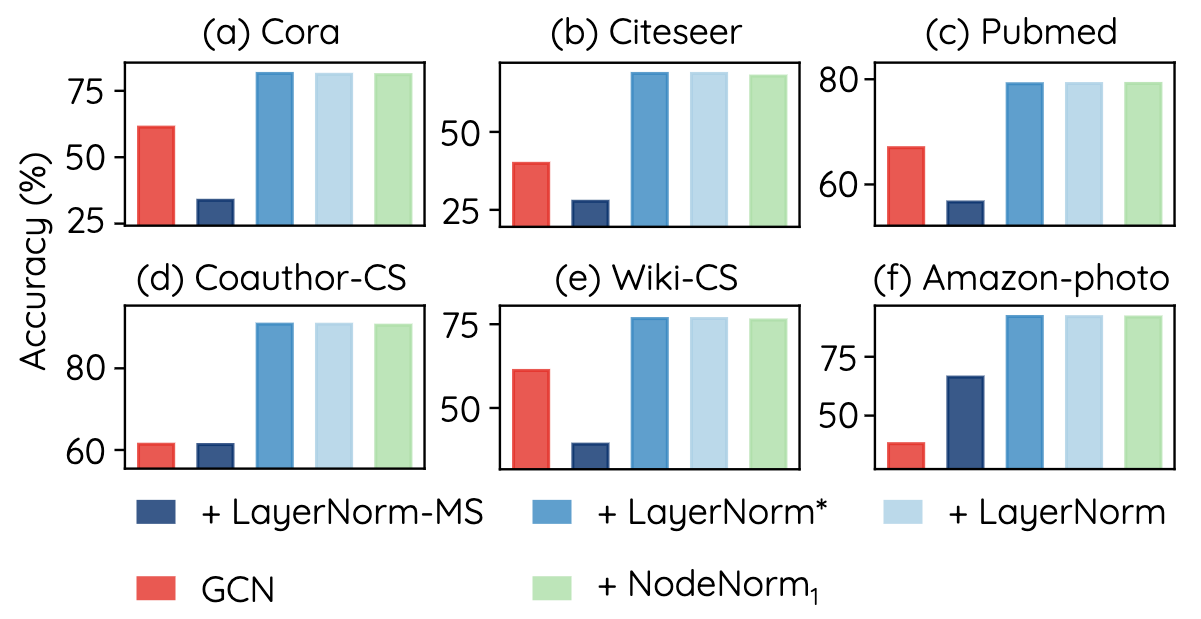}
    \caption{Classification accuracy of 64-layer GCNs with LayerNorm-MS and LayerNorm*. We also include results of the baseline GCNs, GCNs with \nodenormone or LayerNorm for clearer comparison.}
    \label{fig:layernorm-ms}
\end{figure}

We conduct experiments with the two variants of 64 layers, with the same settings as in Sec.~\ref{subsec:address-performance-degradation}, and show results in Fig.~\ref{fig:layernorm-ms}.
We can see that GCNs with LayerNorm*, LayerNorm or \nodenormone perform comparatively, while those with LayerNorm-MS perform significantly worse (on 4 datasets even worse than baseline GCNs).
Note that \nodenormone is equivalent to the variance-scaling step in LayerNorm. The above observations show that linear transformation and the mean-subtraction step are not critical for improving deep GCNs performance; instead, variance-scaling is the step that really works. This demonstrates that reducing \varianceinflaming is the key to addressing performance degradation. 

\section{Conclusion}
\label{conclusion}
In this paper, we investigate the performance degradation problem of GCNs by focusing on effects of TRANsformation operations~(\trans).
We find \trans contribute significantly to the declined performance, providing a new understanding for the community.
Furthermore, we find \trans tend to amplify the node-wise feature variance of node representations, and then as GCNs get deeper, its node-wise feature variance becomes increasingly larger. 
We also find deep GCNs perform significantly worse than shallow ones on nodes with relatively large feature variance.
We thus hypothesize and experimentally justify that mitigating \varianceinflaming effectively addresses performance degradation of GCNs. 
In particular, a simple variance-controlling technique termed \nodenorm, which normalizes each node's hidden features with its own standard deviation, is developed. 
We experimentally prove it can enable deep GCNs (\eg 64-layer) to compete with and even outperform shallow ones (\eg 2-layer).
\nodenorm outperforms existing best methods on addressing performance degradation of GCNs, and can generalize to other GNN architectures.

\bibliographystyle{ACM-Reference-Format}
\balance 
\bibliography{ref} 
\end{document}